\documentclass{article}

\usepackage{microtype}
\usepackage{graphicx}
\usepackage{subcaption}
\usepackage{booktabs} 

\usepackage{hyperref}

\usepackage[preprint]{icml2026}

\usepackage{amsmath}
\usepackage{amssymb}
\usepackage{mathtools}
\usepackage{amsthm}

\usepackage{url}
\usepackage{wrapfig}
\usepackage{algorithm}     
\usepackage{algorithmic}
\usepackage{multirow} 

\usepackage{fancyhdr}
\usepackage{lipsum}
\fancypagestyle{myfootnote}{%
  \fancyhf{}                              
  \renewcommand{\headrulewidth}{0pt}      
  \fancyfoot[L]{%
    \parbox{\textwidth}{%
      \footnotesize \textit{Note:} Several icons in Figure~\ref{fig:intro},~\ref{fig:pipeline} and~\ref{fig:amps_pipe} were generated by OpenAI's ChatGPT, which are solely for illustrative purposes.%
    }%
  }
  \fancyfoot[R]{\thepage}               
}

\usepackage[capitalize,noabbrev]{cleveref}

\theoremstyle{plain}

\theoremstyle{definition}

\theoremstyle{remark}

\usepackage[textsize=tiny]{todonotes}

\icmltitlerunning{AMPS: Adaptive Modality Preference Steering via Functional Entropy}

\begin{document}

\twocolumn[
  \icmltitle{AMPS: Adaptive Modality Preference Steering via Functional Entropy}



  \icmlsetsymbol{equal}{*}

  \begin{icmlauthorlist}
    \icmlauthor{Zihan Huang}{ucsd}
    \icmlauthor{Xintong Li}{ucsd}
    \icmlauthor{Rohan Surana}{ucsd}
    \icmlauthor{Tong Yu}{adobe}
    \icmlauthor{Rui Wang}{adobe}\\
    \icmlauthor{Julian McAuley}{ucsd}
    \icmlauthor{Jingbo Shang}{ucsd}
    \icmlauthor{Junda Wu}{ucsd}
  \end{icmlauthorlist}

  \icmlaffiliation{ucsd}{University of California, San Diego}
  \icmlaffiliation{adobe}{Adobe Research}

  \icmlcorrespondingauthor{Junda Wu}{juw069@ucsd.edu}

  \icmlkeywords{Multimodality, Preference Steering}

  \vskip 0.3in
]



\printAffiliationsAndNotice{}  

\begin{abstract}
Multimodal Large Language Models (MLLMs) often exhibit significant modality preference, which is a tendency to favor one modality over another. 
Depending on the input, they may over-rely on linguistic priors relative to visual evidence, or conversely over-attend to visually salient but facts in textual contexts.
Prior work has applied a uniform steering intensity to adjust the modality preference of MLLMs. 
However, strong steering can impair standard inference and increase error rates, whereas weak steering is often ineffective.
In addition, because steering sensitivity varies substantially across multimodal instances, a single global strength is difficult to calibrate.
To address this limitation with minimal disruption to inference, 
we introduce an instance-aware diagnostic metric that quantifies each modality’s information contribution and reveals sample-specific susceptibility to steering.
Building on these insights, we propose a scaling strategy that reduces steering for sensitive samples and a learnable module that infers scaling patterns,
enabling instance-aware control of modality preference.
Experimental results show that our instance-aware steering outperforms conventional steering in modulating modality preference, achieving effective adjustment while keeping generation error rates low.
\end{abstract}

\section{Introduction}

\begin{figure}[ht]
    \centering
    \includegraphics[width=.82\linewidth]{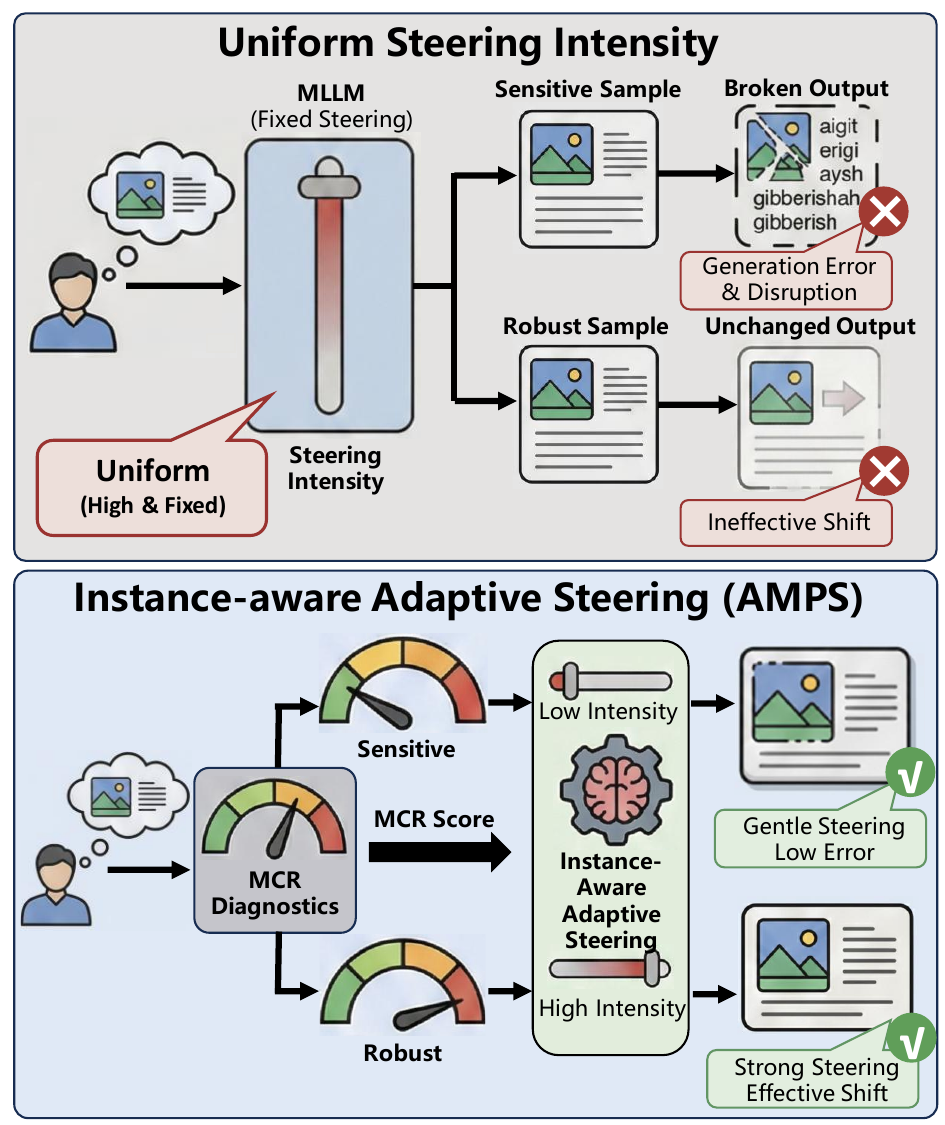}
    \caption{AMPS handles failure cases in modality preference steering~\citep{parekh2025learning} with uniform steering intensity.}
    \label{fig:intro}
    \vspace{-1.5em}
\end{figure}

MLLMs have demonstrated remarkable capabilities in processing and integrating different modalities \cite{huang2025traceable},
such as GUI~\cite{nguyen2025gui,wang2025weakly}, music~\cite{wang2025csymr,mundada2025wildscore}, and audio~\cite{surana2025musicrs,wu2024futga}, enabling complex tasks like visual question answering~\cite{wang2026scenealign,wu2024visual}, multimodal recommendation~\cite{huang2025towards,wu2024personalized}, and cross-modal reasoning~\citep{liu2023visual,wu2025doc,li2022blip}. 
However, the MLLMs exhibit significant modality preference and exhibit an inherent tendency to favor certain modalities~\citep{li2024commit,zheng2025mllms}, 
leading to inconsistent cross-modal reliance.
Depending on the input, they may overweight linguistic priors relative to visual evidence~\citep{liu2025steeringmultimodallargelanguage, zhang2025evaluatingsteeringmodalitypreferences}, or conversely over-attend to visually salient but facts in textual contexts~\citep{zhang2025evaluatingsteeringmodalitypreferences}.
This calls for controllable methods for adjusting MLLMs' modality preferences~\citep{li2024commit, liu2025steeringmultimodallargelanguage, nguyen2026atlasadaptivetesttimelatent}.

Recent works apply steering methods to adjust the modality preferences of MLLMs~\citep{parekh2025learning, ding2025mllmeraserachievingtesttimeunlearning}. 
These methods typically involve introducing calibrated steering vectors to shift the model's activations towards the target modality~\citep{parekh2025learning}.
However, existing steering methods typically fix a global steering strength and apply it uniformly to all input samples~\citep{zhang2025evaluatingsteeringmodalitypreferences, ferrando2025dynamicallyscaledactivationsteering}, as shown in the upper panel of Figure~\ref{fig:intro}. 
Such a strategy is limited because an MLLM’s susceptibility to steering, characterized by how steering strength trades off preference shift against inference disruption, 
depends strongly on the input context.
Applying an excessively strong steering signal to a sample that is inherently sensitive may disrupt the model's standard inference process, 
leading to a high rate of generation errors and degraded performance~\citep{zhang2025evaluatingsteeringmodalitypreferences}.
Conversely, for samples that are robust to changes, a weak steering signal may prove insufficient to elicit any meaningful shift in modality preference, 
rendering the intervention ineffective. This lack of adaptation thus poses a significant challenge to the practical deployment of steering methods. 

A natural way to address this challenge is to adapt steering strength at the instance level, but doing so requires a sample-wise diagnostic signal that quantifies an MLLM’s susceptibility to steering~\citep{wu2025automating, liu2026visionlanguageintrospectionmitigatingoverconfident, ferrando2025dynamicallyscaledactivationsteering}.
To this end, we introduce the Modality Contribution Score (MCS), a diagnostic metric that estimates each modality’s relative information contribution~\citep{cai2025diagnosingmitigatingmodalityinterference, park2024assessingmodalitybiasvideo} during the model’s reasoning process.
MCS enables us to characterize instance-specific susceptibility by revealing systematic variation across inputs in how steering strength trades off modality-preference shift against inference disruption.

Building on this diagnostic signal, we develop an instance-aware steering framework that modulates modality preference while minimizing inference disruption, as shown in the the lower panel of Figure~\ref{fig:intro}.
Specifically, we first propose a steering scaling strategy that sets the steering strength as a function of MCS, using the estimated modality contributions as a proxy for instance-specific susceptibility. 
Intuitively, when an instance is predicted to be highly sensitive to steering, we down-scale the steering strength to avoid destabilizing generation. 
When an instance is predicted to be more robust, we allow stronger steering to induce a meaningful preference shift. 
Thus, we introduce \textbf{Adaptive Modality Preference Steering (AMPS)}, a learnable scaling module that infers the appropriate steering strength directly from the input context and applies it automatically. 
This module is trained to automatically infer and apply appropriate, instance-aware scaling patterns directly from the input, 
enabling a more nuanced and data-driven adjustment of steering intensity. 
Our approach effectively mitigates the limitations of uniform steering, paving the way for more reliable and practical deployment of steering methods in MLLMs. 
The main contributions of our work can be summarized as follows:
\begin{itemize}
    \item We introduce Modality Contribution Score (MCS), an instance-wise diagnostic metric that quantifies per-modality information contribution and predicts susceptibility to steering-induced disruption.
    \item We propose an instance-aware scaling strategy that adjusts steering strength based on MCS to balance preference shift and inference stability.
    \item We develop AMPS, a learnable scaling module that infers steering intensity from input context, and demonstrate consistent gains over prior steering methods.
    \item Our extensive experiments on 8 diverse tasks of the $MC^2$ benchmark show AMPS achieves superior preference steering while reducing generation collapse compared to uniform steering baselines.
\end{itemize}

\section{Preliminary}

\begin{figure*}[ht]
    \centering
    \begin{subfigure}[b]{0.32\textwidth}
        \centering
        \includegraphics[width=\textwidth]{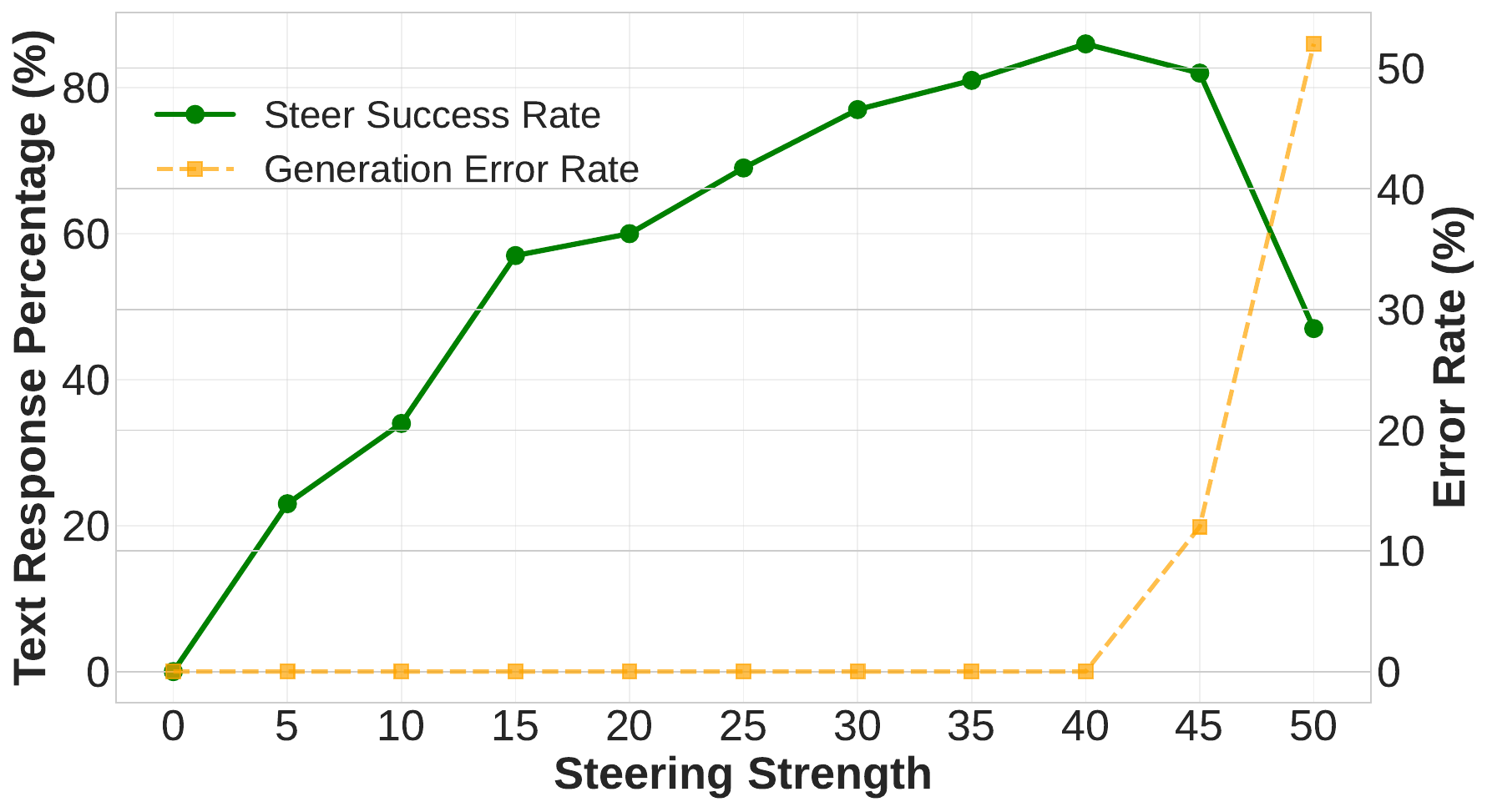}
        \caption{Visual-to-text Meaning Steering}
        \label{fig:visual_to_text}
    \end{subfigure}
    \hfill 
    \begin{subfigure}[b]{0.33\textwidth}
        \centering
        \includegraphics[width=\textwidth]{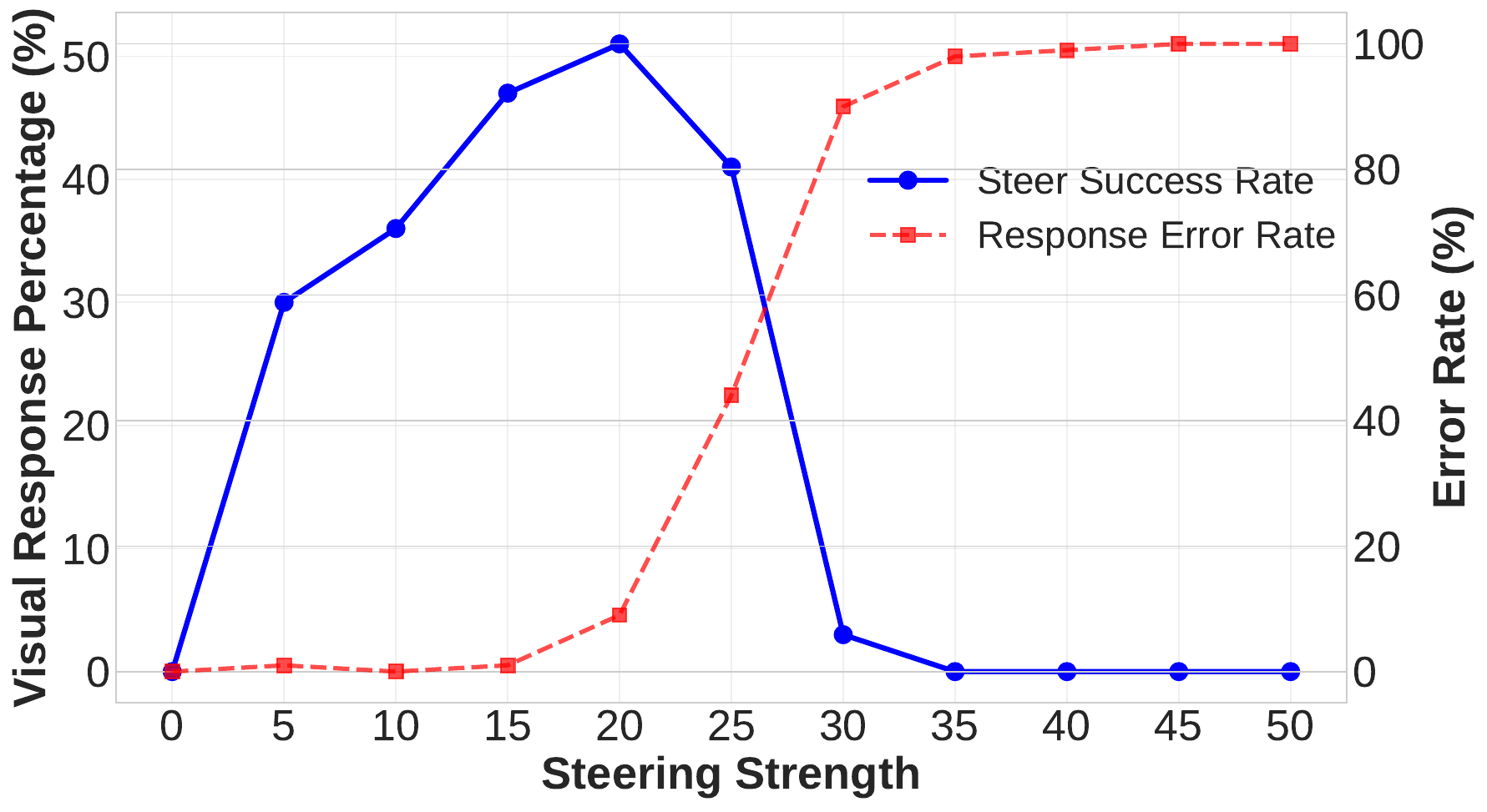}
        \caption{Text-to-visual Meaning Steering}
        \label{fig:text_to_visual}
    \end{subfigure}
    \hfill 
    \begin{subfigure}[b]{0.32\textwidth}
        \centering
        \includegraphics[width=\textwidth]{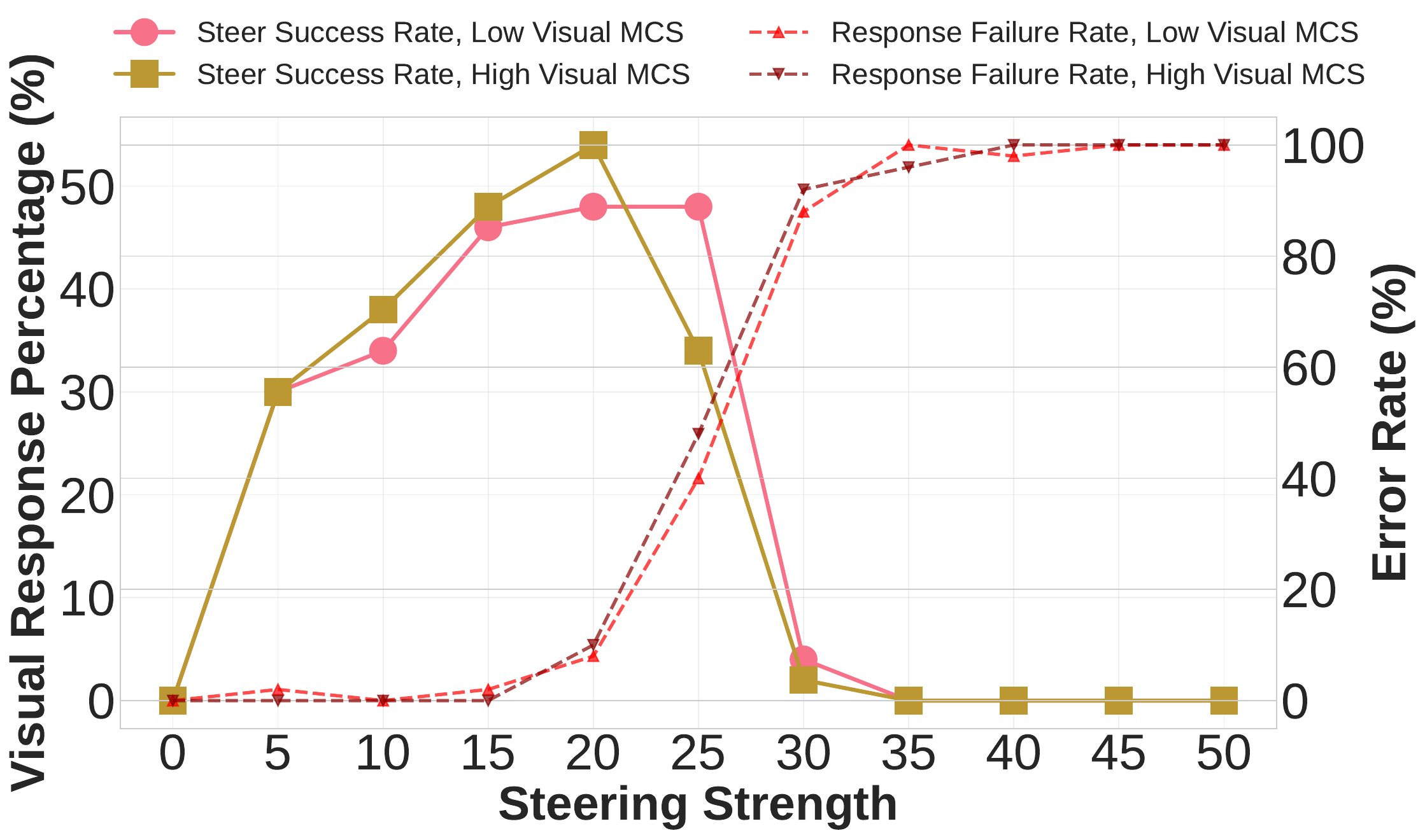}
        \caption{Visual-to-text Meaning Steering Sample Classified with MCS}
        \label{fig:motivation}
    \end{subfigure}
    
    \caption{Mean Steering with different steering intensity.}
    \label{fig:combined_results}
    \vspace{-1.5em}
\end{figure*}

\subsection{MLLM Inference Formulation}

Let $\mathbf{x} = \{\mathbf{T}, \mathbf{V}, \mathbf{P}\}$ denote the multimodal input to a MLLM, where $\mathbf{T}$: Text tokens (instructions/questions) $\mathbf{V}$: Visual tokens (encoded image patches) $\mathbf{P}$: Prompt tokens that trigger modality preference.

Given a MLLM $p_\theta(\cdot)$ parameterized by $\theta$, the visual input $\mathbf{V}$ is first encoded by a vision encoder $e(\cdot)$ such as Vision Transformer and a multimodal projector, producing visual tokens $e(\mathbf{V})$. These visual tokens are concatenated with text tokens $\mathbf{T}$ and optional prompt tokens $\mathbf{P}$, then fed into the large language model backbone $g_\pi$:

\vspace{-\baselineskip} 
\begin{equation}
    \left(\mathbf{h}^{(l)}\right)_{l=1}^L = g_{\pi}^{(l)}\left(e(\mathbf{V}), \mathbf{T}, \mathbf{P}]\right)_{l=1}^L,
\end{equation}
\vspace{-\baselineskip}

where $L$ is the total number of layers, $\pi^{(l)}$ denotes the $l$-th transformer layer, and $\mathbf{h}^{(l)}$ represents the hidden states at layer $l$.

The final layer output is projected via a logit projection layer $\phi$ to produce the output distribution:

\vspace{-\baselineskip} 
\begin{equation}
p_\theta(y_t \mid \mathbf{x}) = \mathrm{softmax}\left(\phi(\mathbf{h}_t^{(L)})\right)
\label{eq:output_logits}
\end{equation}
\vspace{-\baselineskip} 

\subsection{Modality-Specific Functional Entropy of MLLMs}

Functional entropy is defined for continuous random variables. Consider a non-negative function $f(\mathbf{z})$ where $\mathbf{z} \in \mathbb{R}^{d}$ is a stochastic variable with probability measure $\mu$. The functional entropy is defined as:

\vspace{-\baselineskip} 

\begin{equation}
\begin{aligned}
\mathrm{Ent}_{\mu}(f) \triangleq 
& \int_{\mathbb{R}^{d}} f(\mathbf{z})\log f(\mathbf{z}) d\mu(\mathbf{z}) \\
& - \left(\int_{\mathbb{R}^{d}} f(\mathbf{z}) d\mu(\mathbf{z})\right) 
   \log\left(\int_{\mathbb{R}^{d}} f(\mathbf{z}) d\mu(\mathbf{z})\right).
\end{aligned}
\label{eq:entropy_def}
\end{equation}
\vspace{-\baselineskip}

Functional entropy $\mathrm{Ent}_{\mu}(f)$ quantifies the variability of a function $f$ under measure $\mu$. In our setting, $f$ can be used to measure the prediction shift under modality perturbations (Eq.~\ref{eq:ce}). A higher entropy indicates a greater sensitivity to that modality, implying that modality has stronger contribution to influence the output. 

Direct empirical estimation of $\mathrm{Ent}_{\mu}(f)$ is challenging due to the logarithmic integral term. Since $\mathrm{Ent}_{\mu}(f)$ is non-negative, i.e., $\mathrm{Ent}_{\mu}(f) \geq 0$ (with equality if and only if $f(\mathbf{z})$ is constant), it is appropriate to employ the logarithmic Sobolev inequality for Gaussian measures, which bounds the functional entropy by the functional Fisher information:

\vspace{-\baselineskip} 
\begin{equation}
\mathrm{Ent}_{\mu}(f) \leq \frac{1}{2} \int_{\mathbb{R}^{d}} \frac{\|\nabla f(\mathbf{z})\|^{2}}{f(\mathbf{z})} d\mu(\mathbf{z}).    
\label{eq:scaling}
\end{equation}
\vspace{-\baselineskip}

In our case, we define the following function to measure the sensitivity of the MLLMs output logits in Eq~\ref{eq:output_logits} to Gaussian perturbations z on modality-specific hidden states as:

\begin{equation}
    f_M = \mathrm{CE}\left( p_\theta(\cdot \mid \mathbf{z_m}), p_\theta(\cdot \mid \mathbf{x}) \right)
    \label{eq:ce}
\end{equation}
\vspace{-\baselineskip} 

where $m \in \{\mathbf{T}, \mathbf{V}\}$ denotes the perturbed modality, $\mu_m$ is a Gaussian measure centered at $\mathbf{x}_m$ with variance $\sigma_m^2$, $\mathrm{CE}(p,q)$ denotes the cross-entropy between distributions $p$ and $q$, and $\mathbf{z_m}$ represents the input $\mathbf{x}$, after being encoded as hidden representation in KV cache, with K,V value related to modality $m$ perturbed. Specifically, denote modality $m$ related token indexes in both input $\mathbf{x}$ and KV cache as $\mathcal{R}[m]$. For example, if $m$ is visual modality, then $\mathcal{R}[m]$ means the indexes of visual tokens in $\mathbf{z_m}$ (same as image token indexes in $\mathbf{x}$), we add gaussion noise to the K,V values on indexes $\mathcal{R}[m]$ and let MLLM do the forward after this perturbation.

\subsection{Empirical Study of Stability-Efficiency Trade-off in Instance-aware Adaptive Steering}\label{sec:emperical}

A key challenge in steering modality preference in MLLMs is balancing the strength of preference shift against generation stability. Existing methods apply a uniform steering strength to all samples, overlooking sample-dependent sensitivity to intervention. To illustrate this, we conduct a controlled study on 100 samples where Qwen2.5VL-7B initially favors either text or vision. We apply a fixed-direction steering vector to shift preference toward the opposite modality while gradually increasing its intensity.

Results show that moderate steering effectively alters preference, but beyond a threshold, generation quality drops sharply: outputs become nonsensical, repetitive, or empty, as shown in Figure~\ref{fig:visual_to_text}. This highlights a clear stability–efficiency trade-off: strong steering disrupts sensitive samples, while weak steering under-adjusts robust ones.

Further analysis in Figure~\ref{fig:text_to_visual} reveals that susceptibility varies significantly across samples. For instance, inputs with high cross-modal conflict degrade more easily. These observations motivate an instance-aware approach that adapts steering strength based on each sample’s sensitivity. In the following sections, we introduce a diagnostic metric for quantifying this sensitivity and a learnable module for dynamic intensity scaling.

\section{Method}

Building on the empirical observation in section~\ref{sec:emperical}, we propose an adaptive steering framework designed to balance steering efficacy with generation stability. Our goal is to improve the modality preference steering mechanism, focusing on its ability to precisely and robustly shift model bias on a per-instance basis, rather than optimizing for general task performance across diverse downstream applications. The proposed framework consists of: (1) a diagnostic metric, the Modality Contribution Ratio (MCR, section\ref{sec:mcr}), which quantifies modality contribution and sample-wise steering sensitivity; and (2) a learnable Adaptive Modality Preference Steering (AMPS) module (section\ref{sec:steer}), which dynamically scales intervention intensity based on MCR in a context-aware manner, thereby mitigating the stability-efficiency trade-off of uniform steering.

\subsection{Modality Contribution Ratio}
\label{sec:mcr}

To measure modality specific functional entropy of MLLMs, we apply the log-Sobolev inequality shown in Eq~\ref{eq:scaling} with f as Eq~\ref{eq:ce}, which gives a bounding of functional entropy with functional Fisher information, indicating the information contribution from different modalities for MLLMs.

\vspace{-\baselineskip} 
\begin{equation}
\mathrm{Ent}_{\mu}(f_M) \leq \int \frac{ \| \nabla_{\mathbf{z}_m} \text{CE}(p_\theta(\cdot\mid\mathbf{z}), p_\theta(\cdot\mid\mathbf{x})) \|^2 }{ \text{CE}(p_\theta(\cdot\mid\mathbf{z}), p_\theta(\cdot\mid\mathbf{x})) } d\mu_m(\mathbf{z}) \label{eq:sensitivity_bound}  
\end{equation}
\vspace{-\baselineskip} 

The $p_\theta(\cdot\mid\mathbf{x})$ and $p_\theta(\cdot\mid\mathbf{z})$ is the generation using KV cache $KV_{\text{prev}}$ and $KV_{\text{z}}$  as in Algorithm~\ref{alg:sensitivity_estimation} respectively. To facilitate calculation, we apply Monte Carlo sampling to estimate the integral of the right hand side of Equation Equation~\ref{eq:sensitivity_bound}. We firstly denote modality contribution score \textbf{(MCS)} in Equation~\ref{eq:sensitivity}. To estimate the integral, we do $N$ Gaussian perturbations $\mathbf{z}_{m,j} \sim \mu_m$ and calculate Equation~\ref{eq:sensitivity}, which are then used to calculate an average value as the approximation, as in Equation~\ref{eq:mcr_def}.

\vspace{-\baselineskip} 
\begin{equation}
\mathcal{C}_{m,j} =  \frac{ \| \nabla_{\mathbf{z}_{m,j}} \text{CE}(p_\theta(\cdot\mid\mathbf{z}_{j}), p_\theta(\cdot\mid\mathbf{x})) \|^2 }{ \text{CE}(p_\theta(\cdot\mid\mathbf{z}_{j}), p_\theta(\cdot\mid\mathbf{x}))}\label{eq:sensitivity}
\end{equation}
\vspace{-\baselineskip}

To empirically estimate the \textbf{MCS} $\mathcal{C}_m$ derived from our theoretical framework in MLLMs, we implement a KV cache based perturbation gradient analysis on the MLLMs during its forward pass. To obtain a normalized measure of relative modality reliance, we define the \textbf{Modality Contribution Ratio (MCR)} for each modality $m$ as the proportion of its contribution score relative to the total contribution across all modalities in Equation~\ref{eq:mcr_def}.

\begin{equation}
\mathcal{C}_m =  \frac{1}{N} \sum_{j=1}^N \mathcal{C}_{m,j} \text{  ,   }\mathcal{R}_m =  \frac{\mathcal{C}_{m}}{\sum_{m=1}^N \mathcal{C}_{m}}
\label{eq:mcr_def}
\end{equation}

\begin{algorithm}[t]
\caption{Modality Contribution Metric Implementation } 
\label{alg:sensitivity_estimation}
\begin{algorithmic}[1]
\REQUIRE Logits Model $L_\theta$ (Model $p_\theta = \text{softmax}(L_\theta)$) , input $\mathbf{x}$, modality token ranges $\mathcal{R}$, KV cache $KV_{\text{prev}}$, perturbed KV cache $KV_{\text{z}}$ , perturbation strengths $\mathcal{E}$, the Monte Carlo sampling Times $N$. 
\ENSURE MCR $\mathcal{R}_m$ for modalities $m \in \{\text{V}, \text{T}\}$
\STATE $\text{\textbf{Eq.\ref{eq:output_logits}:} } \text{logits}_x, KV_x \gets L_\theta(\mathbf{x}, KV_{\text{prev}})$\\ $\mathbf{P}_x \gets \text{softmax}(\text{logits}_x)$
\FOR{$i \in \{1, \ldots, N\}$}
    \FOR{$m \in \{\text{V}, \text{T}\}$}
        \STATE \textbf{Sample noise:} $\epsilon \sim \mathcal{N}(0, \sigma^2_m \mathbf{I})$
        \STATE \textbf{Perturb KV cache:} \\$KV_z \gets KV_{\text{prev}}$, $KV_z[\mathcal{R}[m]] \gets KV_z[\mathcal{R}[m]] + \epsilon$
        \STATE \textbf{Forward pass:} $\mathbf{P}_z \gets \text{softmax}(L_\theta(\mathbf{x}, KV_z))$
        \STATE \textbf{Compute CE loss (Eq.\ref{eq:ce}):} $\mathcal{L} \gets \text{CE}(\mathbf{P}_x, \mathbf{P}_z)$
        \STATE \textbf{Compute Gradient norm (Eq.\ref{eq:sensitivity}):} \\ $\mathcal{C}_{m,i} \gets \|\nabla_{KV_z} \mathcal{L}\|_2^2 / \mathcal{L}$
    \ENDFOR
\ENDFOR
\STATE \textbf{Compute MCR (Eq.\ref{eq:mcr_def}):}\\ $\mathcal{C}_m \gets \mathbb{E}_{m}[\mathcal{C}_{m,i}]$ for each $m$
\\$\mathcal{R}_m \gets  \frac{\mathcal{C}_{m}}{\sum_{m=1}^N \mathcal{C}_{m}}$ for each $m$
\end{algorithmic}

\end{algorithm}

\begin{figure*}
    \centering
    \includegraphics[width=0.8\linewidth]{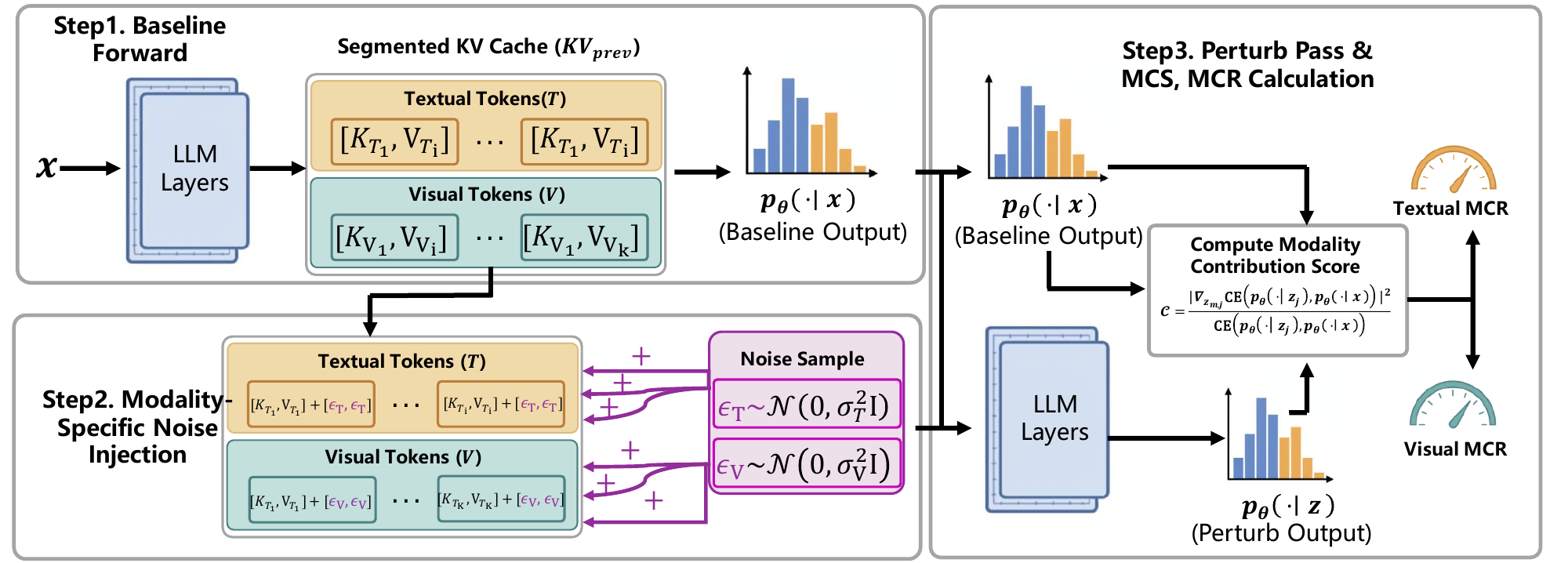}
    \caption{MCR Measurement Pipeline, in this figure the $K_{T_i}$ and $K_{V_j}$ means the corresponding to the index of the i-th text or the index of the j-th visual token in K cache; $V_{T_i}$ and $V_{V_j}$ means the corresponding to the index of the i-th text or index of the j-th visual token in V cache.}
    \label{fig:pipeline}
    \vspace{-1em}
\end{figure*}

The procedure of MCR calculation in MLLMs is outlined in Algorithm~\ref{alg:sensitivity_estimation} and illustrated in Figure~\ref{fig:pipeline}. Specifically, for a given input sequence, we firstly perform a standard forward pass to obtain the original output probability distribution from the MLLMs output logits $p_\theta(\cdot|\mathbf{x})$. Then for each modality $m$ (Visual \textbf{V}, Text \textbf{T}), we introduce isotropic Gaussian noise $\epsilon \sim \mathcal{N}(0, \sigma^2\mathbf{I})$ specifically to the segments of the KV cache corresponding to the target modality. A subsequent forward pass with the perturbed cache yields a new distribution $p_\theta(\cdot|\mathbf{z})$. The contribution score is calculated by computing the squared norm of the gradient of the cross-entropy loss between the original and perturbed distributions with respect to the perturbed KV cache and normalized by the loss itself. This score, averaged across $N$ perturbations, provides a robust empirical measure of the functional Fisher information and, by extension, the model's reliance on each modality.

As in Algorithm~\ref{alg:sensitivity_estimation}, the computational cost of our MCR measurement is $O(N \cdot M \cdot (T_f + T_b))$, where $N$ and $M$ denote the number of Monte Carlo samples and modalities respectively, and $T_f$ and $T_b$ represent the forward and backward pass times. In practice, with $N=3$ and $M=2$, this translates to a $\sim$4.3$\times$ time overhead during training and negligible overhead during inference. Detailed derivation and empirical measurements are provided in Appendix~\ref{sec:comp-cost-details}.

\subsection{Instance-aware Adaptive Steering}
\label{sec:steer}

\begin{figure*}[ht]
    \vspace{-0.5em}
    \centering
    \includegraphics[width=.77\linewidth]{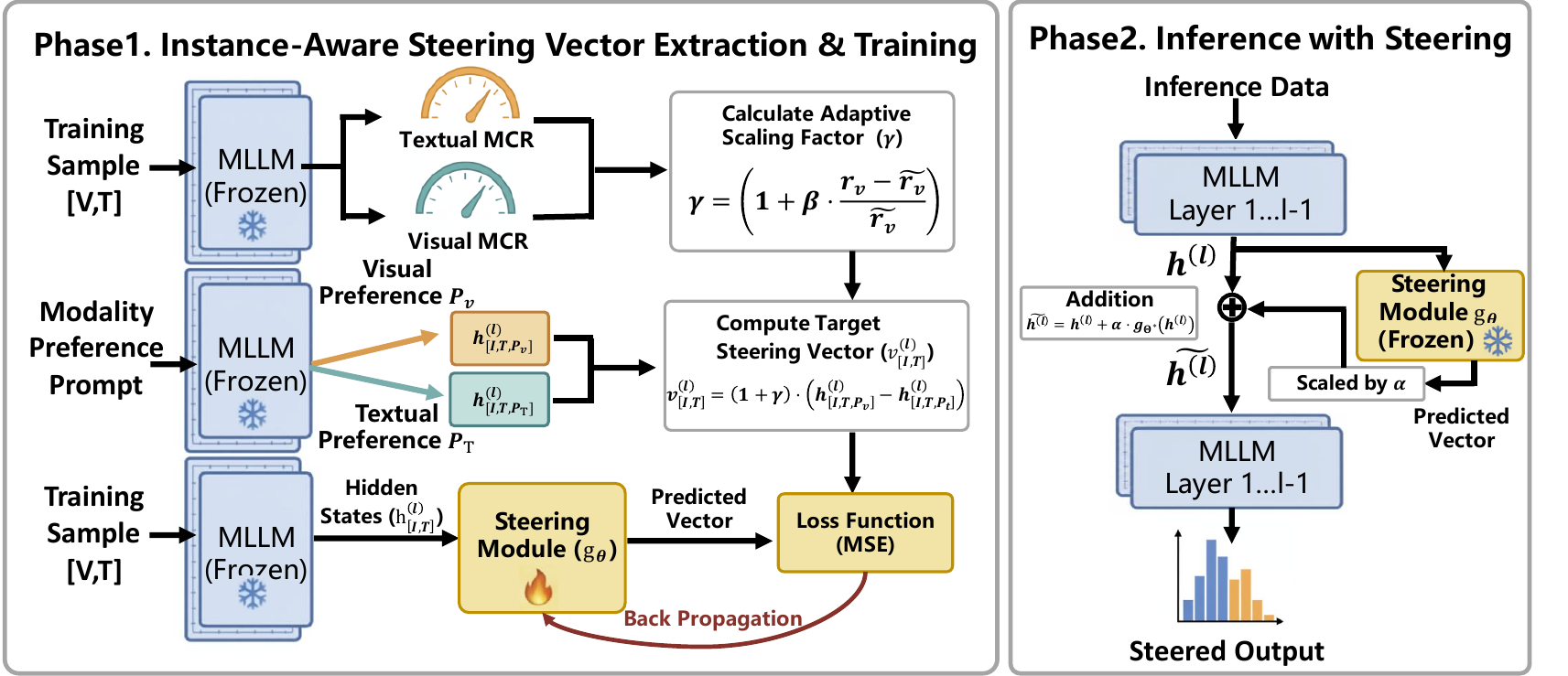}
    \caption{AMPS Training and Application Pipeline}
    \label{fig:amps_pipe}
    \vspace{-1.5em}
\end{figure*}

Conventional steering methods for preference modification construct contrastive pairs by appending prompts such as $\mathcal{P}^{+}$ = ``based on visual context'' and $\mathcal{P}^{-}$ = ``based on text context'' to steer the modality bias in a desired direction:

\vspace{-\baselineskip} 
\begin{equation}
X^{+} = (V, T \parallel P^{+}), \quad X^{-} = (V, T \parallel P^{-})    
\end{equation}
\vspace{-\baselineskip} 

However, directly applying steering based on these steering vector extracted by such pair construction fails to provide a context-aware steering intensity scaling for preference adjustment. The key challenge is a lack of diagnostics of the severity of the model's preference deviation toward expected direction when processing each sample and adjust the intensity of steering in a fine-grained way. Previously studies~\citep{} and our observation in Figure~\ref{fig:motivation} both show steering uniformly with sufficient steering intensity can impair MLLMs' normal inference, which leads to failure in MLLM generation. On the other hand, insufficient steering does not impact MLLMs behaviour sufficiently. Based on our observation that MLLMs exhibit varying modality biases depending on different context, finding a context-aware scaling steering mechanism to mitigate modality bias with different intensity is a prospective solution.

\textbf{Instance-aware Adaptive Steering Vector Extraction.} To bridge this gap, we provide a context-aware steering mechanism, which learn to generate steering vector with different steering intensity based on the severity of MLLMs' modality preference, as shown in Figure~\ref{fig:amps_pipe}. Specifically, during extraction of the steering vector, we save the visual and textual MCR of MLLMs, and then take the average of visual MCR across all samples in the same task as an anchor ratio $\widetilde{r_v}$, since MLLMs does not completely fail on whole tasks, we use the deviation of visual ratio on each sample $r_v$ to the $\widetilde{r_v}$ of each sample to decide its severity of modality preference and use this to scale steering intensity.

\vspace{-\baselineskip} 
\begin{equation}
\gamma_{} = \left(1 + \beta \cdot \frac{r_v - \widetilde{r_v}}{\widetilde{r_v}}\right)     
\label{eq:scale}
\end{equation}
\vspace{-\baselineskip} 

where $\mu_v$ is the anchor visual ratio, $r_v$ is the current visual ratio, $h_l$ represent the hidden state at the index of $|I, T, P_v|$ and $|I, T, P_t|$ respectively, and $\beta$ is used to adjust the context-aware intensity.

Following previous learn-to-steer setting, we extract the hidden states at a layer L of the MLLM on each constructed data pair, and subtract them to form a targeted steering vector. As for the input of learn-to-steer module, we extract the hidden states corresponding to the input image and query without any modality preference prompts.

\vspace{-\baselineskip} 
\begin{equation}
\begin{aligned}
h_{[I,T,P_v]}^{(l)} &= g_\pi^{(l)}\left([e(V),T,P_v]\right), \\
h_{[I,T,P_t]}^{(l)} &= g_\pi^{(l)}\left([e(V),T,P_t]\right), \\
h_{[I,T]}^{(l)} &= g_\pi^{(l)}\left([e(V),T]\right).
\end{aligned}
\label{eq:hidden_sub}    
\end{equation}
\vspace{-\baselineskip} 

Combine Eq~\ref{eq:scale} and Eq~\ref{eq:hidden_sub}, we apply the scaling which form the predict target of the steering module.

\vspace{-\baselineskip} 
\begin{equation}
v_{[I,T]}^{(l)} = \left(1 + \gamma\right) \cdot \left( h_{[I,T,P_v]}^{(l)} - h_{[I,T,P_t]}^{(l)}\right)  
\label{eq:v_target}
\end{equation}
\vspace{-\baselineskip} 

\textbf{Learning to Predict Steering Vectors.} The goal is to train a learn-to-steer module $g_\Theta$ that predicts $v^{(l)}$ in Eq~\ref{eq:v_target} from $h_{[I,T]}^{(l)}$ as the third term of Eq~\ref{eq:hidden_sub}, with an optimization objective formulated as:

\vspace{-\baselineskip} 
\begin{equation}
\Theta^* = \arg\min_{\Theta} \mathbb{E}_{[I,T]} \left\| v_{[I,T]}^{(l)} - g_\Theta\left(h_{[I,T]}^{(l)}\right) \right\|_2^2 
\end{equation}
\vspace{-\baselineskip} 

We follow conventional learn-to-steer to construct $g_\Theta$ with a light-weight 2-layer MLP as the auxiliary network and train the network using all the samples we collected. The details of steering module training is illustrated in Figure~\ref{fig:amps_pipe}.

\textbf{Steering Modality Preference in MLLM.} As shown in Figure~\ref{fig:amps_pipe}, we apply the light-weight trained module on one layer's hidden states of MLLMs which predict a scaled steering vector and apply the steering vector back to the hidden state of MLLMs at this layer during MLLMs inference, following conventional l2s experiment setting, we take the 14-th layer in our experiment.

\vspace{-\baselineskip} 
\begin{equation}
\tilde{h}^{(l)} = h^{(l)} + \alpha \cdot g_{\Theta^*}\left(h^{(l)}\right)
\end{equation}
\vspace{-\baselineskip} 

where $\tilde{h}^{(l)}$ is the steered hidden state at layer $l$, $h^{(l)}$ is the original hidden state, and $\alpha$ is a scaling hyperparameter controlling the intensity of steering.

\begin{figure*}[htp]
    \centering
    \begin{minipage}[b]{0.64\linewidth} 
        \centering
        \includegraphics[width=\linewidth]{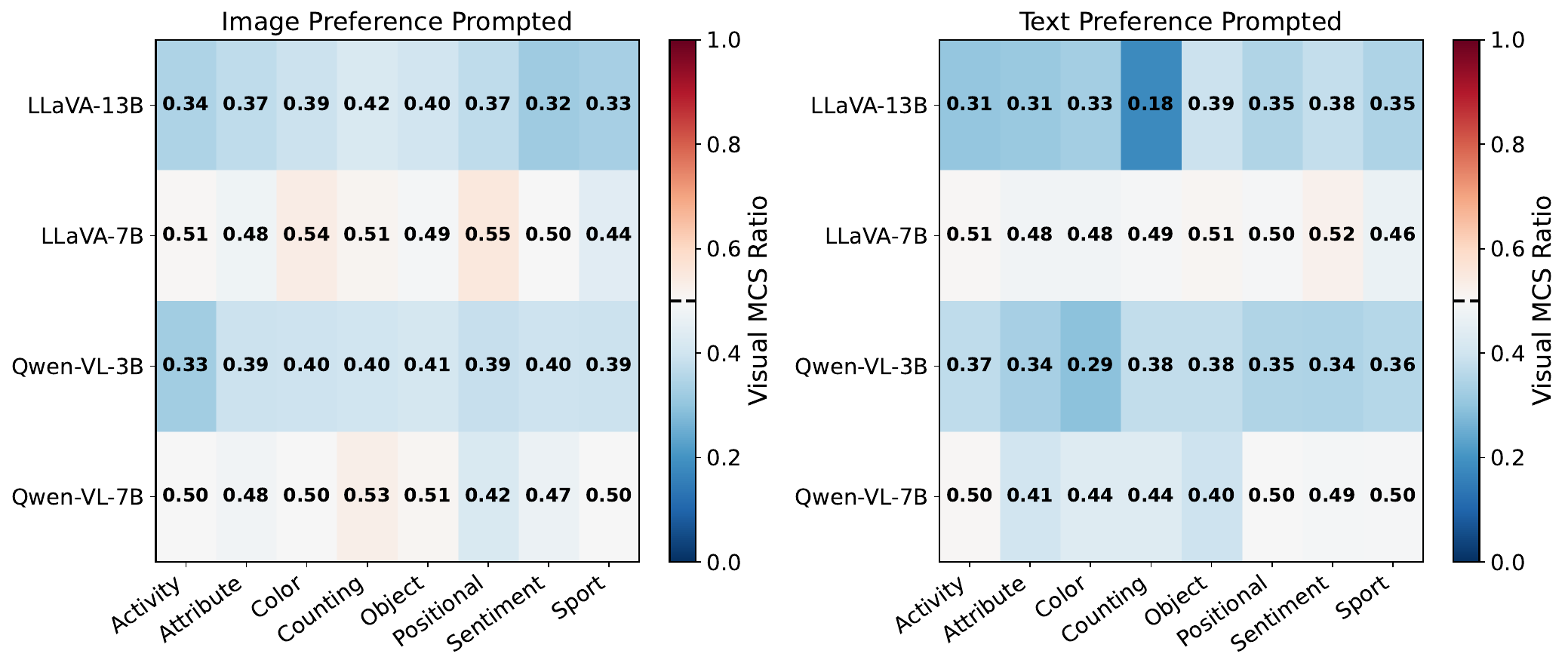}
        \caption{Sensitivity scores of different models across different tasks.}
        \label{fig:sensitivity}
    \end{minipage}%
    \begin{minipage}[b]{0.26\linewidth} 
        \centering
        \includegraphics[width=\linewidth]{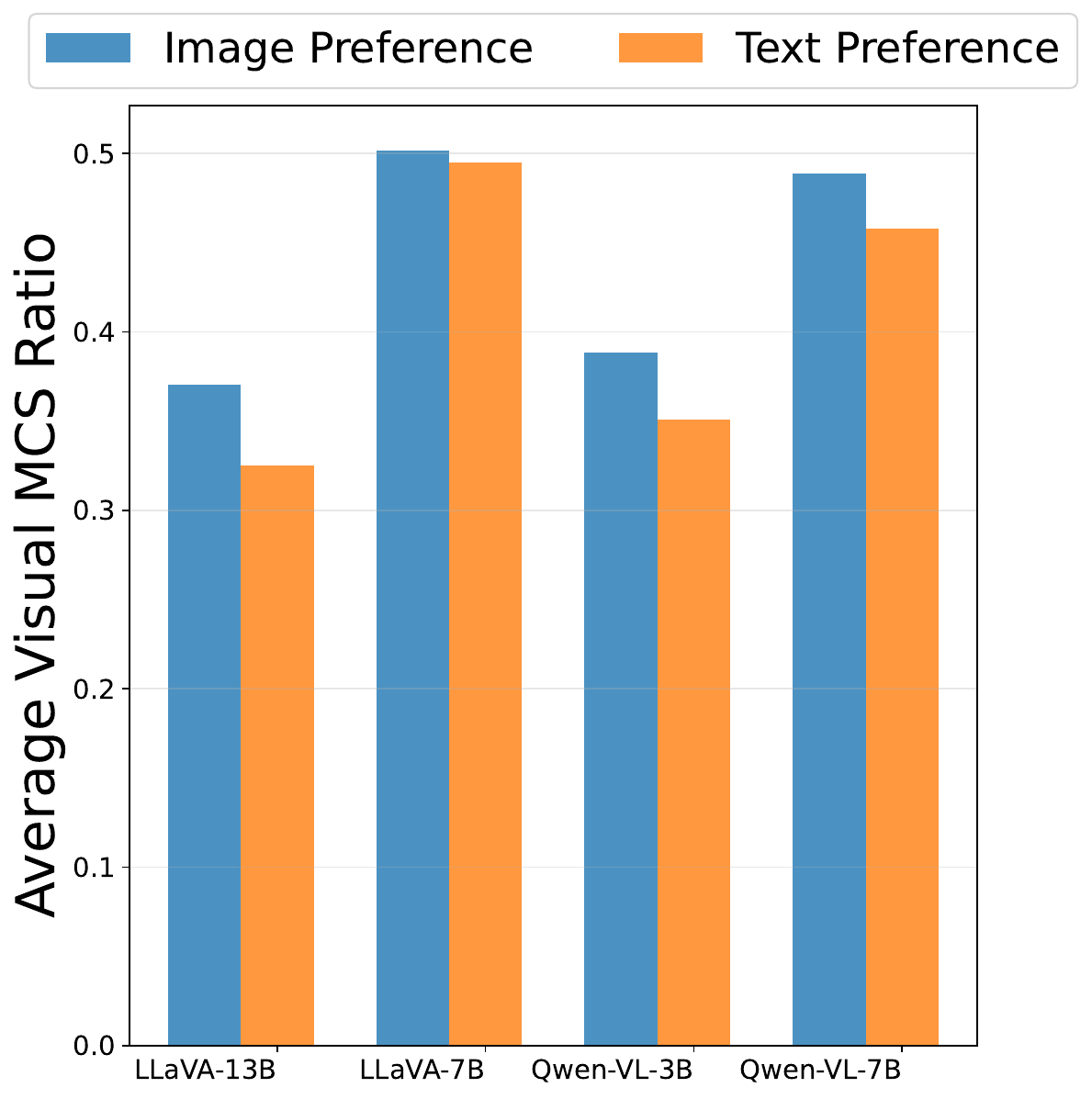}
        \caption{Visual MCS with preference prompts.}
        \label{fig:comparison}
    \end{minipage}
    \label{fig:combined_metric}
    \vspace{-1em}
\end{figure*}

\begin{table*}[ht]
\centering
\small
\caption{Performance comparison with conventional steering and preference modification methods}
\label{tab:performance_1}

\resizebox{0.85\linewidth}{!}{

\begin{tabular}{lccccccccc}
\toprule
\multirow{2}{*}{Preference Model} & \multicolumn{8}{c}{Attribute} & \multirow{2}{*}{Total} \\
\cmidrule(lr){2-9}
 & Sport & Sentiment & Positional & Counting & Color & Activity & Object & Sentiment & \\
\midrule
\textbf{Vision} & & & & & & & & & \\
MLLM-only & 26.4 & 12.4 & 0.8 & 13.2 & 4.0 & 16.0 & 11.6 & 38.0 & 15.3 \\
Inst Design & 60.8 & 24.0 & 20.0 & 20.4 & 10.8 & 32.0 & 27.2 & 63.2 & 32.3 \\
CoT Prompting & 57.6 & 27.6 & 16.0 & 18.0 & 21.6 & 35.6 & 44.4 & 52.4 & 34.2 \\
Few shot & 32.0 & 12.0 & 10.0 & 11.0 & 2.0 & 25.0 & 9.0 & 38.0 & 17.2 \\
Qwen2VL-7B & 78.8 & 35.6 & 38.8 & 29.6 & 22.4 & 56.4 & 45.2 & 78.4 & 48.1 \\
Qwen2.5VL-7B-AMPS & \textbf{67.6} & \textbf{68.0} & \textbf{85.6} & \textbf{66.8} & \textbf{88.8} & \textbf{71.2} & \textbf{52.8} & \textbf{83.2} & \textbf{73.00} \\
\midrule
\textbf{Text} & & & & & & & & & \\
MLLM-only & 12.8 & 46.0 & 68.8 & 38.0 & 39.6 & 20.0 & 43.6 & 14.0 & 35.4 \\
Inst Design & 14.4 & 46.8 & 72.8 & 40.4 & 55.6 & 24.4 & 35.6 & 11.6 & 37.7 \\
CoT Prompting & 27.2 & 63.6 & 83.6 & 62.8 & 75.6 & 53.2 & 58.0 & 20.4 & 55.6 \\
Few shot & 21.0 & 77.0 & 89.0 & 73.0 & 73.0 & 60.0 & 70.0 & 42.0 & 63.1 \\
Qwen2.5VL-7B & 69.6 & 67.6 & 84.4 & 50.8 & 82.8 & 57.6 & 54.8 & 41.2 & 63.6 \\
Qwen2.5VL-7B-AMPS & \textbf{83.2} & \textbf{80.8} & \textbf{68.0} & \textbf{77.6} & \textbf{50.8} & \textbf{82.0} & \textbf{96.8} & \textbf{70.4} & \textbf{76.2} \\
\bottomrule
\end{tabular}
}
\vspace{-0.5em}
\end{table*}

\section{Experiment}

\begin{table*}[ht]
\centering
\small
\caption{Adjusting MLLMs' modality preference on $MC^2$~\cite{zhang2025evaluatingsteeringmodalitypreferences} dataset toward \textbf{Visual}.}
\label{tab:comparison_l2s_visual}
\resizebox{0.85\linewidth}{!}{
\begin{tabular}{@{}l*{12}{c}@{}}
\toprule
\multirow{3}{*}{Task}
& \multicolumn{3}{c}{LLaVA-7B} & \multicolumn{3}{c}{LLaVA-13B} & \multicolumn{3}{c}{Qwen-VL-3B} & \multicolumn{3}{c}{Qwen-VL-7B} \\
\cmidrule(l){2-4}\cmidrule(l){5-7}\cmidrule(l){8-10}\cmidrule(l){11-13}
& Base & l2s & AMPS & Base & l2s & AMPS & Base & l2s & AMPS & Base & l2s & AMPS \\
\midrule
Act.   & 12.4 & 18.0 & \textbf{22.0} & 10.0 & 18.0 & \textbf{47.2} & 28.0 & 30.8 & \textbf{44.0} & 29.6 & 46.0 & \textbf{67.6} \\
Attr.  &  8.0 & 11.6 & \textbf{18.4} &  8.4 & 10.0 & \textbf{32.8} & 24.0 & 24.4 & \textbf{28.4} & 28.0 & 53.6 & \textbf{68.0} \\
Color  & 12.0 & 13.6 & \textbf{27.6} &  8.0 & 14.8 & \textbf{54.0} & 41.2 & 55.2 & \textbf{61.6} & 58.8 & 76.4 & \textbf{85.6} \\
Count. &  2.8 &  6.0 & \textbf{12.8} &  2.4 &  6.8 & \textbf{49.2} & 12.0 & 17.2 & \textbf{32.0} & 31.2 & 43.2 & \textbf{66.8} \\
Obj.   & 33.2 & 36.0 & \textbf{43.2} & 28.0 & 39.6 & \textbf{66.4} & 60.8 & 68.4 & \textbf{76.0} & 72.4 & 76.8 & \textbf{88.8} \\
Pos.   & 22.8 & 16.0 & \textbf{28.8} & 14.8 & 18.0 & \textbf{52.4} & 31.6 & 40.0 & \textbf{42.8} & 33.2 & 46.8 & \textbf{71.2} \\
Sent.  &  0.4 &  1.6 & \textbf{ 2.4} &  2.0 &  4.0 & \textbf{16.4} &  6.8 &  6.4 & \textbf{12.4} &  8.8 & 19.6 & \textbf{52.8} \\
Sport  & 22.0 & \textbf{25.6} & 23.6 & 28.0 & 44.8 & \textbf{78.8} & 64.0 & 70.4 & \textbf{75.6} & 65.6 & 69.6 & \textbf{83.2} \\ \midrule
Total  & 14.2 & 16.05 & \textbf{22.35} & 12.70 & 19.50 & \textbf{49.65} & 33.55 & 39.10 & \textbf{46.6} & 42.00 & 54.0 & \textbf{73.00} \\
\bottomrule
\end{tabular}
}
\end{table*}

\begin{table*}[ht]
\centering
\small
\caption{Adjusting MLLMs' modality preference on $MC^2$~\cite{zhang2025evaluatingsteeringmodalitypreferences} dataset toward \textbf{Text}.}
\label{tab:comparison_l2s_text}
\resizebox{0.85\linewidth}{!}{
\begin{tabular}{@{}l*{12}{c}@{}}
\toprule
\multirow{3}{*}{Task}
& \multicolumn{3}{c}{LLaVA-7B} & \multicolumn{3}{c}{LLaVA-13B} & \multicolumn{3}{c}{Qwen-VL-3B} & \multicolumn{3}{c}{Qwen-VL-7B} \\
\cmidrule(l){2-4}\cmidrule(l){5-7}\cmidrule(l){8-10}\cmidrule(l){11-13}
& Base & l2s & AMPS & Base & l2s & AMPS & Base & l2s & AMPS & Base & l2s & AMPS \\
\midrule
Act.   & 82.0 & 78.4 & \textbf{86.0} & 89.6 & 91.6 & \textbf{92.8} & 72.0 & 75.2 & \textbf{86.8} & 70.4 & 86.8 & \textbf{87.2} \\
Attr.  & 91.6 & 91.2 & \textbf{92.0} & 92.0 & 93.6 & \textbf{95.6} & 76.0 & 82.0 & \textbf{86.0} & 72.0 & \textbf{83.2} & 82.8 \\
Color  & 87.2 & 85.2 & \textbf{92.0} & 86.8 & 88.4 & \textbf{90.0} & 58.8 & 70.0 & \textbf{76.8} & 41.2 & 68.4 & \textbf{71.6} \\
Count. & 87.6 & 87.6 & \textbf{94.0} & 96.4 & \textbf{96.8} & 96.0 & 88.0 & 92.8 & \textbf{94.8} & 68.4 & 77.2 & \textbf{83.2} \\
Obj.   & 62.8 & \textbf{78.0} & 65.2 & 69.2 & \textbf{82.8} & 81.6 & 39.2 & 44.8 & \textbf{54.0} & 26.0 & \textbf{64.8} & 64.0 \\
Pos.   & 86.0 & 85.6 & \textbf{88.0} & 83.2 & 83.2 & 85.2 & 68.4 & 75.6 & \textbf{84.0} & 66.8 & 79.6 & \textbf{83.2} \\
Sent.  & 90.8 & 90.8 & \textbf{92.8} & 90.0 & 90.4 & \textbf{92.0} & 93.2 & 96.0 & \textbf{98.0} & 90.4 & \textbf{96.0} & 94.6 \\
Sport  & 74.8 & 75.6 & \textbf{76.0} & 74.4 & \textbf{86.0} & 82.8 & 36.0 & 43.2 & \textbf{54.4} & 34.4 & \textbf{73.6} & 73.2 \\ \midrule
Total  & 82.85 & 84.05 & \textbf{85.75} & 85.20 & 89.10 & \textbf{89.50} & 66.45 & 72.45 & \textbf{79.35} & 57.60 & 78.7 & \textbf{80.2} \\
\bottomrule
\end{tabular}
}
\end{table*}

The primary focus of our experiment is to investigate and improve the modality preference steering mechanism in MLLMs with diagnostic metrics specifically designed for analyzing modality reliance and steering dynamics, rather than enhancing general downstream task performance. Our results reflect the capability of AMPS to precisely adjust modality preference while preserving generation stability.

\textbf{Experiment Setup.} We conduct experiments on the Modality Context Conflict dataset ($\mathbf{MC}^2$) \citep{zhang2025evaluatingsteeringmodalitypreferences}, a benchmark derived from real-world image-text distributions sourced from TDIUC~\citep{DBLP:conf/iccv/KafleK17} and MS-COCO~\citep{DBLP:conf/eccv/LinMBHPRDZ14}. $\mathbf{MC}^2$ has explicit image-text conflict annotations to quantify modality preference and evaluate steering methods under controlled yet ecologically valid scenarios. We compare AMPS against key baselines: \textit{prompt engineering} \textit{static steering}, and the state-of-the-art adaptive controller \textit{learn-to-steer (L2S)} \citep{parekh2025learning}. All experiments freeze the base MLLM parameters and are performed on Qwen-2.5VL~\citep{Qwen-VL} (3B, 7B) and LLaVA1.5~\citep{liu2023improvedllava} (7B, 13B) to ensure robustness across scales and architectures. Further details regarding dataset composition, baseline selection rationale, and hyperparameter configurations are provided in Appendix \ref{app:experiment_setup}.

\begin{table*}[ht]
\centering
\small
\caption{Ablation study on modality preference adjustment for Qwen2.5VL-3B on $MC^2$.}
\label{tab:ablation_study}
    \begin{tabular}{@{}lccccccccc@{}}
    \toprule
    Model & \multicolumn{8}{c}{Task Accuracy (\%)} & Overall \\
    \cmidrule(lr){2-9}
     & Act. & Attr. & Color & Count. & Obj. & Pos. & Sent. & Sport & Total \\
    \midrule
    Qwen-VL-3B (w/o steering) & 28.0 & 24.0 & 41.2 & 12.0 & 60.8 & 31.6 & 6.8 & 64.0 & 33.55 \\
    Qwen-VL-3B-l2s (w/o scaling) & 30.8 & 24.4 & 55.2 & 17.2 & 68.4 & 40.0 & 6.4 & 70.4 & 39.10 \\
    Qwen-VL-3B-AMPS (full) & \textbf{44.0} & \textbf{28.4} & \textbf{61.6} & \textbf{32.0} & \textbf{76.0} & \textbf{42.8} & \textbf{12.4} & \textbf{75.6} & \textbf{46.60} \\
    \bottomrule
    \end{tabular}
\parbox{\linewidth}{\footnotesize \textit{Note:} This table presents the ablation study results. "w/o steering" denotes the baseline model without steering mechanism, "w/o scaling" indicates the model without scaling component, and "full" represents the complete proposed method. Best results are highlighted in bold.}
\end{table*}

\subsection{RQ1: Can the proposed Modality Contribution Score (MCS) effectively quantify and reveal the modality preference patterns in MLLMs?}

Using MCR measurement pipeline, we quantitatively diagnose the modality preference patterns of MLLMs. We compute Visual and Textual MCS values, and visualize the ratio of Visual MCS to the total MCS, on the $MC^2$ dataset.

We conduct this analysis on two model families, Qwen-VL and LLaVA, with different parameter sizes under two controlled settings designed to induce contradictory preferences, \textit{visual preference} and \textit{text preference}. To elicit these preferences, we append specific instructional prompts to the original query. For instance, prompts like \textit{"Respond to the question based only on the image context"} are used to encourage visual preference, while instructions such as \textit{"Follow the text context rather than the image content"} are used to promote text preference. The distribution of the Visual MCS ratio across different tasks is visualized in Figure~\ref{fig:sensitivity}, and the aggregated results are summarized in Figure~\ref{fig:comparison}. Our analysis reveals two key observations:

\textbf{MLLMs exhibit significant variation in modality reliance in different input contexts and are highly sensitive to instructional prompts.} As shown in Figure~\ref{fig:sensitivity}, the Visual MCS ratio varies considerably not only across different tasks but also among different models. This indicates that the relative contribution of visual information to the model's reasoning is not static but is highly context-dependent. Furthermore, this contribution score is systematically and predictably influenced by modality-preference prompts, demonstrating that MLLMs' intrinsic preference can be steered through external instructions.

\textbf{Instructional prompts effectively shift global modality preference, as quantified by the MCS metric.} The aggregated results in Figure~\ref{fig:comparison} show a clear and consistent trend: prompts designed to induce a visual preference lead to a statistically significant increase in the overall Visual MCS, while prompts inducing a text preference result in a correspondingly higher Textual MCS. This confirms that our proposed MCS metric serves as a reliable indicator of the models' internal preference dynamics.

\subsection{RQ2: Does our Adaptive Modality Preference Steering (AMPS) framework outperform existing prompt engineering and static steering methods?}

We first compare the effectiveness of our method against two baselines, prompt engineering and static prompt steering. As shown in Table~\ref{tab:performance_1}, our method applied to Qwen2.5-VL 7B achieves a significantly larger shift in modality preference in both directions (text-to-visual and visual-to-text) compared to baseline methods. This result indicates that our adaptive approach based on learnable module exert a far stronger range of control over modality preference than either heuristic prompt design or steering methods that apply static adjustment.

\begin{figure}[h]
    \centering
    \vspace{-0.5em}
\includegraphics[width=0.85\linewidth]{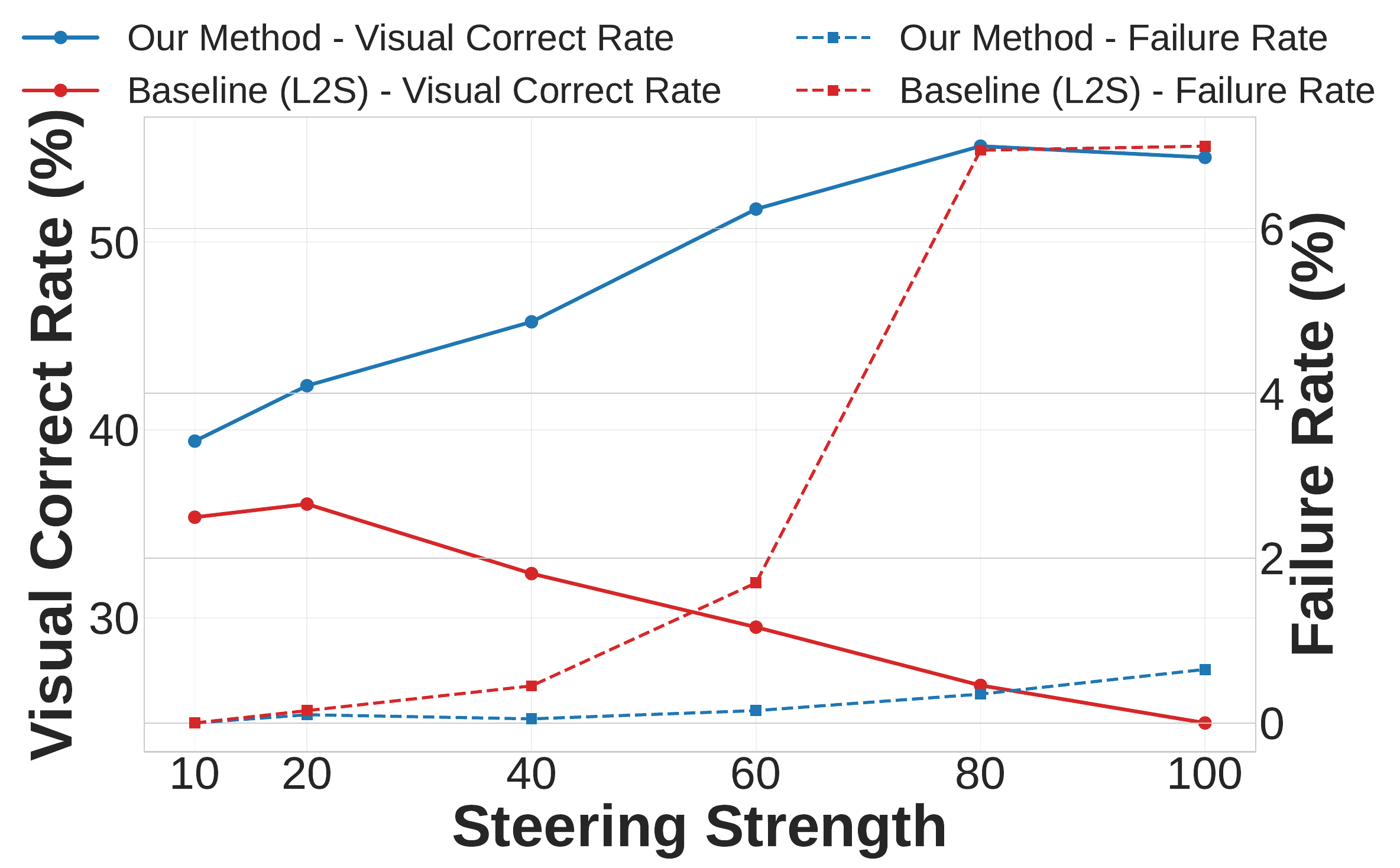}
    \caption{Comparison of AMPS and conventional learn-to-steer on different steering intensity.}
    \label{fig:analysis}
    \vspace{-1em}
\end{figure}

\subsection{RQ3: How does the proposed scaling mechanism contribute to the steering performance?}

To study the benefit of our proposed scaling strategy, we conduct an experiment where we compare the performance of the AMPS with traditional learn-to-steer module without the scaling mechanism. As shown in Table~\ref{tab:comparison_l2s_visual} and \ref{tab:comparison_l2s_text}, we evaluate these variants across different model architectures including Qwen2.5-VL and LLaVA1.5, and different scales from 3B to 13B. Our comparison demonstrate that while the base learn-to-steer module effectively shifts the modality preference, integrating our scaling mechanism yields a consistent and significant performance gain. This improvement is particularly pronounced for the more challenging text-to-visual steering direction, underscoring the benefits of sample-wise adaptation.

\subsection{RQ4: How does the scaling mechanism enhance the steering dynamics and robustness?}

We further dissect how the scaling mechanism enhances steering effectiveness. Figure~\ref{fig:analysis} illustrates the performance of AMPS compared to the conventional learn-to-steer module across a spectrum of steering intensities. The analysis reveals that our method achieves a more effective preference shift across a wider range of intensities while mitigating the performance collapse typically observed when excessive steering intensity is applied to sensitive samples. This demonstrates that our scaling mechanism enables a more stable and reliable steering process.

\subsection{Ablation Study}

We further conduct an ablation study to validate the contribution of each component in our framework, with results presented in Table~\ref{tab:ablation_study}. The study demonstrates that both proposed components are essential for achieving optimal performance. The learn-to-steer module alone ("w/o scaling") provides a significant boost over the baseline without steering, confirming its foundational role in modality preference adjustment. Moreover, incorporating the steering intensity scaling mechanism to form the full model yields a further substantial performance gain across all tasks. This consistent improvement highlights the critical function of scaling, which adapts the steering intensity per sample, thereby maximizing effectiveness and preventing degradation. 

\section{Related Work}

\paragraph{Modality preference in MLLMs.}
MLLMs exhibit stable \emph{modality preferences} under cross-modal conflict \cite{li2024commit,huang2025listwise,huang2025image,huang25traceable}. \citet{zhang2025evaluatingsteeringmodalitypreferences} measure these preferences via MC$^{2}$, showing they are encoded in steerable latent representations. Concurrent analyses report systematic text-leaning effects in vision tasks \citep{deng2025words} and quantify modality importance in video QA \citep{park2025assessing}. Preference is often framed as language-prior dominance, addressed via decoding-time calibration to re-weight visual evidence \citep{zhang2024debiasing}. These studies motivate the diagnostics to make modality reliance observable and manipulable.

\paragraph{Adaptive Steering and Control.}
Representation-level steering adjusts modality reliance post-hoc, e.g., via hidden-space shifts \citep{zhang2025evaluatingsteeringmodalitypreferences,wu2025mitigating} or input-dependent vectors \citep{parekh2025learning}. Other approaches include causal attention interventions \citep{zhou2024mitigating}, adaptive controllers \citep{wu2025automating}, training-time alignment \citep{zhang2025debiasing,jiang2024modality}, decoding-time penalties \citep{zhang2024debiasing}, and model-editing techniques \citep{10.1145/3664647.3681589}. Mechanistically, visual under-utilization is linked to attention distribution issues, such as mid-layer degradation \citep{jiang2025devils} and visual attention sinks \citep{kang2025see}. Decoding-time fixes like VAR, IKOD, iTaD redistribute attention to sustain grounding \citep{kang2025see,yang2025ikodmitigatingvisualattention,xu2025mitigating}.

\section{Conclusion}

This work proposes a sample‑wise diagnostic metric (MCR) and an adaptive steering method to overcome the uniform‑intensity limitation in adjusting MLLMs’ modality preference. The MCR metric quantifies each modality’s relative contribution to reasoning and reveals sample‑level susceptibility to steering. By dynamically scaling steering intensity based on this diagnostic, our approach effectively shifts modality preference while reducing steering‑induced errors. Experiments on the diverse tasks of the $MC^2$ benchmark and across multiple model configurations show consistent improvements over conventional steering methods. Diagnostic analysis confirms that MCR reliably captures intrinsic modality reliance across different input contexts in controlled conflict settings, offering a principled basis for adaptive preference steering. Overall, this work advances robust, context‑aware control over multimodal reasoning and contributes to more trustworthy and interpretable steering of MLLM behavior.

\section{Impact Statement}
This paper presents work whose goal is to advance the field of machine learning. There are many potential societal consequences of our work, none of which we feel must be specifically highlighted here.

\thispagestyle{myfootnote}

\bibliography{iclr2025_conference}
\bibliographystyle{icml2026}

\newpage
\appendix

\section{Appendix:}
\subsection{Experiment Set Up}
\label{app:experiment_setup}

\textbf{Dataset Details.} The $\mathbf{MC}^2$ dataset is derived from real-world image-text distributions sourced from TDIUC \citep{DBLP:conf/iccv/KafleK17} and MS-COCO \citep{DBLP:conf/eccv/LinMBHPRDZ14}. It encompasses 8 diverse task types, including sentiment analysis, object counting, and positional reasoning. While broader benchmarks like MME \citep{fu2025mmecomprehensiveevaluationbenchmark}, MM-Vet \citep{DBLP:conf/icml/YuYLWL0WW24}, and LLaVA-Bench \citep{liu2023visual} are valuable for general evaluation, they lack the structured modality-conflict annotations required for our focused study of preference dynamics and steering failure mechanisms, making $\mathbf{MC}^2$ uniquely suitable.

\textbf{Baseline Rationale.} We compare against prompt engineering, static steering, and L2S \citep{parekh2025learning} as the most relevant adaptive controller for our task. Other adaptive controllers or decoding-time reweighting methods, such as AutoSteer \citep{wu2025automating} (safety-oriented) and CausalMM \citep{DBLP:conf/iclr/ZhouY0WLH25} (causal mediation), are designed for distinct objectives, such as safety and hallucination reduction, and are not directly applicable to explicit modality-conflict resolution.

\textbf{Hyperparameter Configuration.} Key hyperparameters follow established protocols \citep{zhang2025evaluatingsteeringmodalitypreferences}. The configurations of the learn-to-steer module adhere to the original implementation \citep{parekh2025learning}. The perturbation strengths $\mathcal{E}$ are set to 0.1 and 1 times the standard deviation of the KV cache for Qwen and LLaVA models, respectively. The Monte Carlo sample count $N$ for MCR estimation is set to 3, providing a balance between approximation accuracy and computational overhead.

\subsection{Computational Cost Analysis}
\label{sec:comp-cost-details}

\subsubsection{Theoretical Complexity Analysis}
\label{subsec:theoretical-complexity}

Algorithm~\ref{alg:sensitivity_estimation} computes the Modality Contribution Ratio (MCR) via Monte Carlo estimation of the functional Fisher information. Let us analyze its computational cost step by step.

\paragraph{Time Complexity:}
The algorithm's execution time is dominated by two main operations per Monte Carlo sample and modality:
\begin{enumerate}
    \item \textbf{Forward pass with perturbed KV cache}: This involves a complete inference pass through the Transformer model after modifying cached representations.
    \item \textbf{Gradient computation}: Computing the gradient of the cross-entropy loss with respect to the perturbed cache requires backpropagation.
\end{enumerate}
Let $T_f$ denote the time for a standard forward pass and $T_b$ for a backward pass. With $N$ Monte Carlo samples and $M$ modalities, the total time complexity is:
\begin{equation}
    T_{\text{MCS}} = O(N \cdot M \cdot (T_f + T_b)).
\end{equation}
In the worst case, $T_b \approx 2 \cdot T_f$ due to gradient computation overhead, leading to a theoretical $3N \cdot M$ factor relative to a single forward pass. For $N=3$ and $M=2$, this yields a $18\times$ overhead. However, practical implementations achieve significant efficiency gains through:
\begin{itemize}
    \item \textbf{Optimized gradient computation}: Automatic differentiation reuses intermediate activations, reducing $T_b$ to approximately $T_f$.
    \item \textbf{Batched operations}: Multiple perturbations can be processed in parallel when hardware permits.
\end{itemize}
Thus, the practical overhead approaches $N \cdot M = 6\times$, which aligns with our empirical measurements of $4.3\times$.

\paragraph{Memory Complexity:}
The primary memory overhead comes from additional KV cache copies for perturbation. During MCS computation, we maintain:
\begin{itemize}
    \item Original KV cache for the reference forward pass.
    \item Perturbed KV cache copies for gradient computation.
\end{itemize}
Since we process perturbations sequentially, the peak memory usage involves one additional copy of the KV cache beyond standard inference. This explains the modest $2.1\%$ memory increase observed in practice, as the KV cache represents only a fraction of the total model memory footprint.

\subsubsection{Empirical Measurements}
\label{subsec:empirical-measurements}

We provide detailed empirical measurements on Qwen2.5VL-3B (averaged over 20 samples), validating the theoretical analysis and demonstrating practical efficiency.

\paragraph{Time Cost:}
\begin{table}[h]
\centering
\small
\begin{tabular}{lccr}
\toprule
\textbf{Phase} & \textbf{Operation} & \textbf{Time (s)/Sample} & \textbf{Overhead} \\
\midrule
Training & Standard forward & 0.34 & 1.00$\times$ \\
Training & MCS ($N=1$) & 0.37 & 1.09$\times$ \\
Training & MCS ($N=3$) & 1.46 & 4.29$\times$ \\
Inference & Standard & 0.39 & 1.00$\times$ \\
Inference & AMPS-steered & 0.40 & 1.03$\times$ \\
\bottomrule
\end{tabular}
\caption{Time cost measurements on Qwen2.5VL-3B. MCS with $N=3$ introduces $\sim$4.3$\times$ overhead during training, while inference overhead is negligible.}
\label{tab:time-cost}
\end{table}

The measured 4.29$\times$ overhead (vs. theoretical maximum of $6\times$) results from hardware parallelization and computational reuse.

\paragraph{Memory Consumption:}
\begin{table}[h]
\centering
\small
\begin{tabular}{lcrr}
\toprule
\textbf{Scenario} & \textbf{Memory (MB)} & \textbf{Overhead} \\
\midrule
Standard inference & 7797 & --- \\
MCR inference ($N=3$) & 7959 & +162 MB (2.08\%) \\
\midrule
Standard inference & 7717 & --- \\
AMPS-steered inference & 7739 & +22 MB (0.29\%) \\
\bottomrule
\end{tabular}
\caption{Memory consumption measurements. MCS measurement adds $\sim$2\% memory during steering module training, while steering application adds $<$0.3\% during inference.}
\label{tab:memory-cost}
\end{table}

The memory overhead remains modest due to efficient memory management and sequential processing of perturbations.

\subsubsection{Approximations and Their Implications}
\label{subsec:approximations}

Our practical implementation makes several key approximations to balance computational cost and diagnostic fidelity:

\paragraph{Finite Monte Carlo Samples ($N=3$):}
The Monte Carlo estimator variance scales inversely with sample count. We find $N=3$ provides a favorable trade-off between estimation accuracy and computational cost. In practice, the MCS signal exhibits high correlation across independent runs, confirming robustness despite limited samples.

\paragraph{Fixed Perturbation Strength:}
We use predetermined perturbation magnitudes calibrated to each modality's typical activation scale. This fixed approach avoids the computational overhead of adaptive perturbation tuning while maintaining sensitivity across diverse inputs.

\paragraph{Targeted Cache Perturbations:}
Rather than perturbing the entire KV cache, we modify only segments corresponding to specific modalities. This focused approach reduces computational overhead while preserving the ability to distinguish modality contributions.

\subsubsection{Offline vs. Online Computation}
\label{subsec:offline-online}

A crucial design decision is treating MCS measurement as \textit{offline} computation:
\begin{itemize}
    \item \textbf{Training phase}: MCS is computed once per sample during steering vector collection. The $\sim$4.3$\times$ time overhead is acceptable as it's a one-time cost amortized over the training process.
    \item \textbf{Inference phase}: Only the lightweight steering module (2-layer MLP) is applied, adding minimal latency and memory overhead.
\end{itemize}
This separation ensures real-time applicability while maintaining the diagnostic benefits of MCS measurement.

\subsubsection{Summary}
The computational cost of MCR measurement is manageable in practice: $\sim$4.3$\times$ time overhead during training with 3 samples, and $\sim$2\% memory overhead. The strategic approximations (limited samples, fixed perturbation strength, targeted modifications) provide an effective balance between diagnostic fidelity and computational efficiency. During inference, AMPS adds negligible cost, making it suitable for deployment in performance-sensitive applications.

\onecolumn

\end{document}